\documentclass{article}

\usepackage[preprint]{neurips_2026}

\usepackage[utf8]{inputenc} 
\usepackage[T1]{fontenc}    
\usepackage{hyperref}       
\usepackage{url}            
\usepackage{booktabs}       
\usepackage{amsfonts}       
\usepackage{nicefrac}       
\usepackage{microtype}      
\usepackage{xcolor}         
\usepackage{graphicx}

\usepackage{algorithm}
\usepackage{algpseudocode}
\usepackage{amsmath, amssymb}
\usepackage{hyperref}
\usepackage{cleveref}
\usepackage{placeins} 
\usepackage{makecell}
\usepackage{wrapfig}

\newcommand{\methodname}{\textsc{Cats}}

\title{\methodname{}: Cascaded Adaptive Tree Speculation for Memory-Limited LLM Inference Acceleration}

\author{%
  Yuning Han \\
  University of Florida\\
  Gainesville, FL 32611 \\
  \texttt{yuninghan@ufl.edu} \\
  \And
  Yangchenchen Jin \\
  University of Florida\\
  Gainesville, FL 32611 \\
  \texttt{yangchenchen.jin@ufl.edu} \\
  \And
  Dylan Zhao \\
  University of Florida\\
  Gainesville, FL 32611 \\
  \texttt{dylan.zhao@ufl.edu} \\
  \And
  Jingwei Sun \\
  University of Florida\\
  Gainesville, FL 32611 \\
  \texttt{sun.jingwei@ufl.edu} \\
}

\begin{document}

\maketitle

\begin{abstract}
  Auto-regressive decoding in Large Language Models (LLMs) is inherently memory-bound: every generation step requires loading the model weights and intermediate results from memory (e.g., High-Bandwidth Memory (HBM) for GPU servers), making throughput bottlenecked by memory bandwidth rather than compute. Speculative decoding addresses this by enabling parallel verification of multiple draft tokens, effectively amortizing the cost of each target-model call. However, existing speculative decoding methods are designed under the assumption that HBM is sufficiently large to hold both the target model and an auxiliary draft model simultaneously---an assumption that breaks down on memory-constrained devices such as edge platforms with limited DRAM. We analyze the inference bottleneck in this memory-limited regime and propose \methodname{}, a self-speculative decoding framework that conducts cascaded verification and correction based on the memory budget and parameter offloading patterns on memory-limited devices. This design maximizes token acceptance rate and end-to-end speedup while keeping the peak memory footprint on the device equal to that of the target model alone. We evaluate \methodname{} on different models across five benchmarks on real edge devices. \methodname{} can achieve a wall-clock speedup of up to \textbf{5.08$\times$} with no degradation in generation quality, outperforming the SOTA method by up to \textbf{1.45$\times$} under edge memory constraints. Code is available at \url{https://github.com/ElizaFuLan/CATS.git}.
\end{abstract}

\section{Introduction}
\label{introduction}

Deploying Large Language Models (LLMs) for efficient inference remains one of the central challenges in modern machine learning systems. LLM inference must generate tokens one at a time in an autoregressive loop---each step conditioned on all prior outputs and requiring a full model forward pass. This sequential structure is inherently memory-bound: every generation step loads the model weights and intermediate results from memory (e.g., High-Bandwidth Memory (HBM) for GPU servers) to on-chip compute units, and throughput is governed by memory bandwidth rather than arithmetic capacity~\cite{vellaisamy2026taxbreak, alizadeh2024llmflash}. Speculative decoding was proposed to break this bottleneck through \emph{parallel decoding}~\cite{leviathan2023fast, chen2023speculative, stern2018blockwise}: a lightweight draft model speculatively generates a sequence of candidate tokens, which the larger target model then verifies in a single batched forward pass. Because the target model processes multiple tokens simultaneously, the per-token memory bandwidth cost is amortized over several accepted tokens, substantially improving throughput while preserving the exact output distribution.

However, this throughput gain comes at a structural cost: classical speculative decoding introduces a \emph{second} set of weights---the auxiliary draft model---that must coexist with the target model in device memory, inflating both the static memory footprint and the per-step bandwidth traffic. Self-speculative decoding methods~\cite{zhang2024draftverify, elhoushi2024layerskip, xia2025swift, liu2024kangaroo} emerged as a practical response to this added memory cost: by generating drafts from a shallow sub-network of the target model itself, they eliminate the need for a separate draft model. However, they still need to introduce additional adapter parameters for drafting, and the performance of these methods often suffers from the sub-network's limited drafting capacity.

\begin{wrapfigure}{r}{0.40\textwidth}
  \vspace{-4mm}
  \centering
  \includegraphics[width=\linewidth]{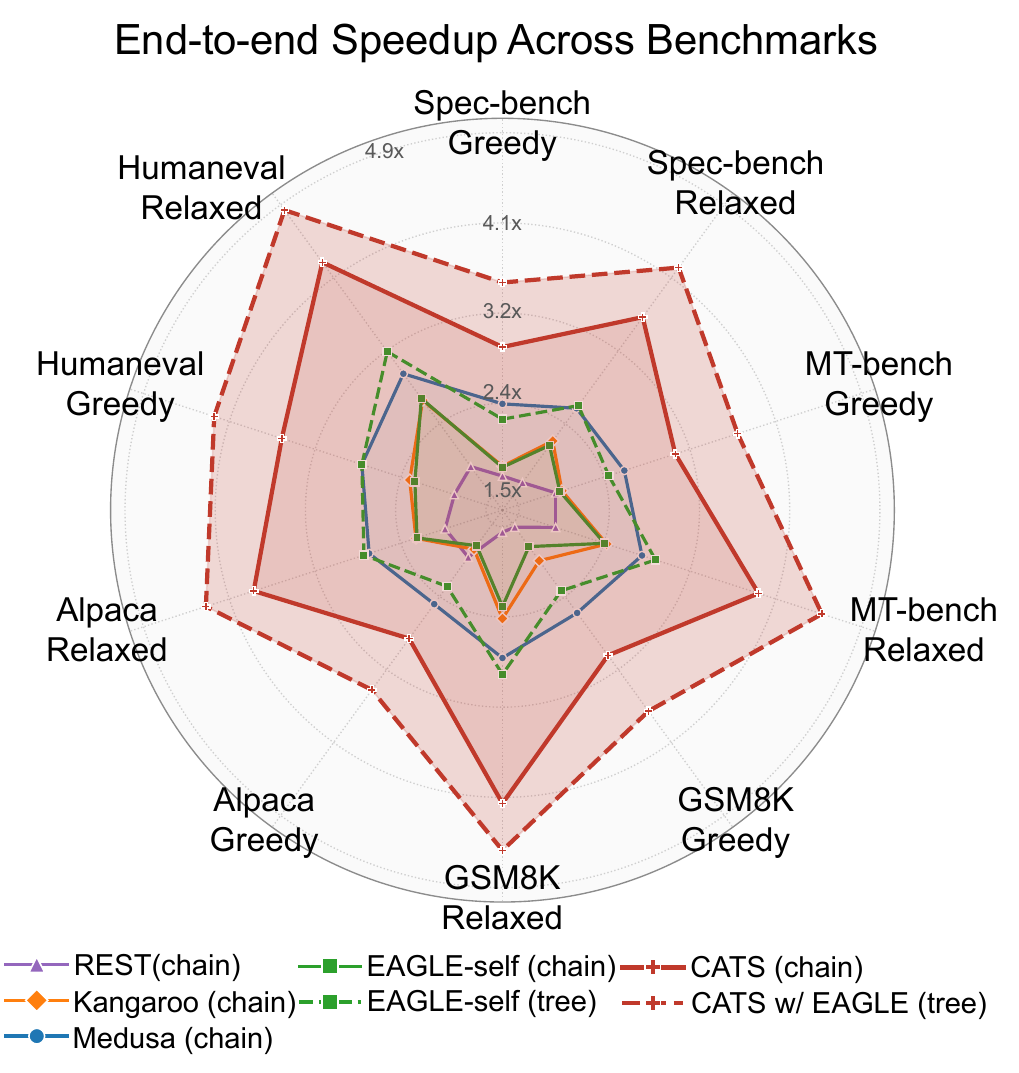}
  \vspace{-4mm}
  \caption{End-to-end speedup across five benchmarks on Vicuna-7B.}
  \label{fig:radar_main_result}
  \vspace{-4mm}
\end{wrapfigure}

A more fundamental limitation runs through \emph{all} existing speculative decoding approaches: they assume model weights remain resident in device memory (e.g., HBM for servers and DRAM for edge devices) for the duration of inference. On large-scale servers, this holds, but on memory-limited platforms such as edge devices, DRAM capacity falls well short of a single 7B-parameter model and must be shared with the operating system---so weights must be streamed from flash memories on every forward pass~\cite{alizadeh2024llmflash}, making flash$\leftrightarrow$DRAM the rate-limiting transfer rather than DRAM$\leftrightarrow$on-chip memory. Every design choice in existing methods---draft depth, verification strategy, acceptance policy---targets the memory-resident regime and is misaligned with this memory-constrained setting. Auxiliary-model approaches are doubly penalized: maintaining a separate draft model adds memory capacity demands and extra transfer traffic at every speculative step, compounding the bottleneck they were meant to relieve.

We investigate this memory-limited inference bottleneck and propose \methodname{} (\textbf{C}ascaded \textbf{A}daptive \textbf{T}ree \textbf{S}peculation), a self-speculative decoding framework that performs cascaded verification to minimize inference time within the memory budget. \methodname{} organizes inference into three stages whose layer boundaries are determined by the available DRAM budget: (1)~a \textbf{draft} stage using a shallow sub-network to draft candidate tokens by repeating draft iterations; (2)~a \textbf{shallow verification} stage using intermediate layers under the DRAM budget, streamed from flash once per cycle, to check draft tokens and produce correction candidates in parallel; and (3)~a \textbf{target verify} stage where the remaining layers are offloaded from the flash chunk by chunk to process a tree-structured input over the draft trace and all correction branches to select the longest accepted prefix. We also propose a \textbf{Reduced KL Loss} that focuses distillation supervision on high-probability tokens to maximize the sub-network's drafting and verification capabilities.

Our main contributions are as follows:
\begin{itemize}
    \item We identify flash$\leftrightarrow$DRAM data movement as the dominant bottleneck for speculative decoding on edge devices---a constraint entirely overlooked by existing methods---and propose \methodname{}, a cascaded self-speculative decoding framework for this memory-constrained regime.
    \item \methodname{} is \textbf{model-agnostic}: the cascaded framework applies to any transformer-based LLM without architectural modifications.
    \item We implement our framework on edge devices. Extensive experiments on Vicuna-7B/13B and LLaMA2-7B/13B across five benchmarks demonstrate that \methodname{} achieves up to \textbf{5.08$\times$} speedup, outperforming all compared self-speculative decoding baselines including \textsc{Kangaroo}~\cite{liu2024kangaroo}, \textsc{Medusa}~\cite{cai2024medusa}, and \textsc{Eagle}~\cite{li2024eagle} under the memory-limited setting.
\end{itemize}

\section{Related Work}
\label{related_work}

\paragraph{Speculative decoding with auxiliary draft models.}
The canonical speculative decoding framework~\cite{leviathan2023fast, chen2023speculative, xia2023specdec} relies on a separately trained smaller model to propose candidates that the target model verifies in parallel. Subsequent work has extended this to tree-structured verification~\cite{miao2024specinfer, chen2024sequoia, wang2024opttree, xiong2024dyspec}, knowledge-distilled draft models~\cite{zhou2024distillspec, liu2024online, du2024glide}, retrieval-based drafting~\cite{he2024rest, li2024nnspec, oliaro2025suffixdecoding}, and lookahead generation~\cite{fu2024lookahead}; see \cite{xia2024survey} for a comprehensive survey. Methods that rely on independently trained smaller models~\cite{kim2023bild, chen2024cascade, zhao2024ouroboros, wang2025specrag, zhong2025sprinter, sun2023spectr, sun2025blockverify, bachmann2025judge, wang2025think, sun2024triforce, chen2025magicdec} incur significant memory overhead, a challenge in edge scenarios where even the target model barely fits in device memory.

\paragraph{Self-speculative decoding.}
To eliminate the dependency on a separate draft model, a line of \emph{self-speculative} methods derives the draft model directly from the target model's own layers. Early-exit approaches such as \textsc{Draft \& Verify}~\cite{zhang2024draftverify} and \textsc{LayerSkip}~\cite{elhoushi2024layerskip} route tokens through only the first few transformer layers for drafting and use the full model for verification. \textsc{Swift}~\cite{xia2025swift} selects skip layers dynamically per input without fine-tuning, while \textsc{KnapSpec}~\cite{cha2026knapspec} formulates layer selection as a knapsack optimization problem. \textsc{Kangaroo}~\cite{liu2024kangaroo} introduces a lightweight adapter on top of the shallow sub-network trained to mimic the target model's output distribution, achieving strong acceptance rates with minimal overhead. Multi-head approaches such as \textsc{Medusa}~\cite{cai2024medusa} and \textsc{Hydra}~\cite{ankner2024hydra} attach parallel prediction heads to the target model's final layer, enabling several future tokens to be proposed in a single forward pass. The \textsc{Eagle} series~\cite{li2024eagle, li2025eagle2, li2025eagle3} drafts at the feature level and constructs dynamic token trees, achieving state-of-the-art speedups on server-class hardware. Consistency-based training~\cite{kou2024cllms, guo2025selfspec} and multi-token prediction objectives~\cite{gloeckle2024mtp, qin2024mjsd, monea2023pass} provide complementary approaches for parallel generation. A common limitation across all these methods is that they still require additional adapter parameters for drafting, which can diminish the performance gains in memory-limited settings achieved by existing speculative decoding methods. And the performance of these methods often suffers from the sub-network's limited drafting capacity.

\paragraph{LLM inference on memory-constrained hardware.}
LLM in a Flash~\cite{alizadeh2024llmflash} characterizes the flash-memory$\leftrightarrow$DRAM bandwidth bottleneck on edge devices where model weights cannot reside in DRAM and proposes windowed loading and row-column bundling to reduce transfer volume per forward pass. PowerInfer~\cite{song2024powerinfer} exploits activation sparsity to selectively load only hot neurons. These works~\cite{ren2025nvme, chen2025edgellm} recognize the memory bottleneck as the critical constraint on edge inference — a constraint that our method seeks to address for existing speculative decoding methods in the same setting.

\section{Preliminary and Motivation}
\label{Motivation}

\begin{figure}[!htbp]
    \centering
    \includegraphics[width=1.0\linewidth]{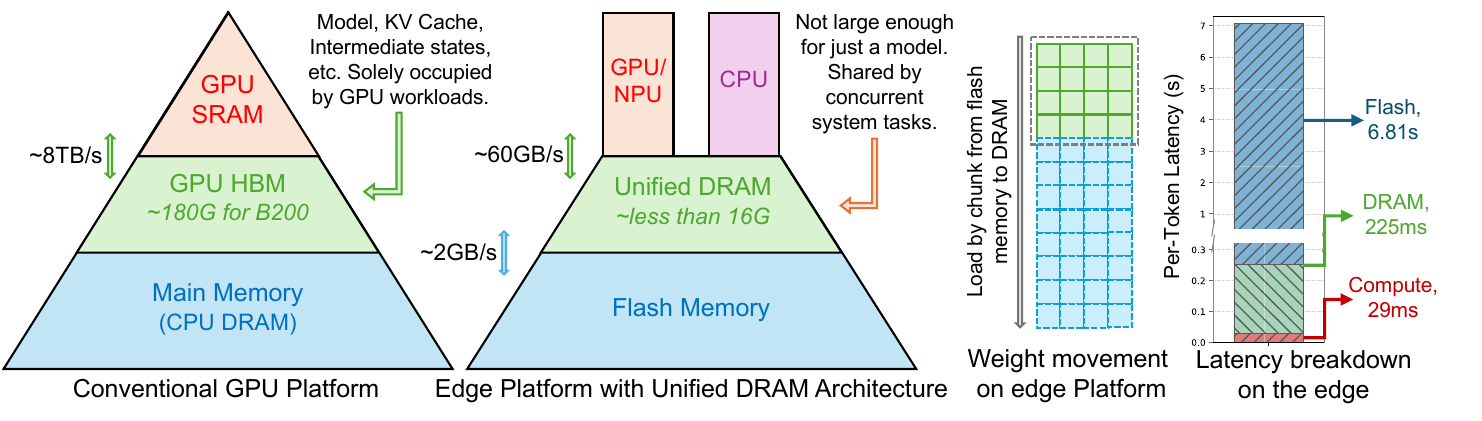}
    \vspace{-1.5em}
    \caption{Memory hierarchy on server vs.\ edge, with measured per-token latency breakdown for autoregressive Vicuna-7B inference. \emph{Server (B200):} model weights reside in HBM; the binding bottleneck is HBM$\leftrightarrow$SRAM bandwidth, and per-token latency is compute-dominated. \emph{Edge (Jetson AGX Orin):} DRAM cannot hold the full model, so weights must be staged from flash on every forward pass; flash$\leftrightarrow$DRAM transfer dominates per-token latency.}
    \label{fig:memory_hierarchy}
\end{figure}


The structural mismatch identified in Section~\ref{introduction}---that all speculative decoding methods assume memory-resident weights, a condition that fails on edge---stems from a fundamental difference in memory hierarchy between the two deployment regimes, which Figure~\ref{fig:memory_hierarchy} makes explicit. On a conventional GPU platform (e.g., NVIDIA B200), HBM provides $\sim$180\,GB at $\sim$8\,TB/s, comfortably holding the target model, an auxiliary draft model, and all intermediate states simultaneously; the binding cost during decoding is the fast HBM$\leftrightarrow$SRAM path. On an edge platform with a unified DRAM architecture, the GPU/NPU and CPU share a single DRAM pool of $\lesssim$16\,GB---difficult to host even a single 7B-parameter model ($\sim$14\,GB), let alone an extra draft model. The data between DRAM and the flash memory moves at only $\sim$2\,GB/s. During the inference, model weights must be loaded chunk by chunk from flash into DRAM on every forward pass~\cite{alizadeh2024llmflash}, and the binding bottleneck

\begin{wrapfigure}{r}{0.40\textwidth}
 \vspace{-4mm}
\centering
     \includegraphics[scale=0.33]{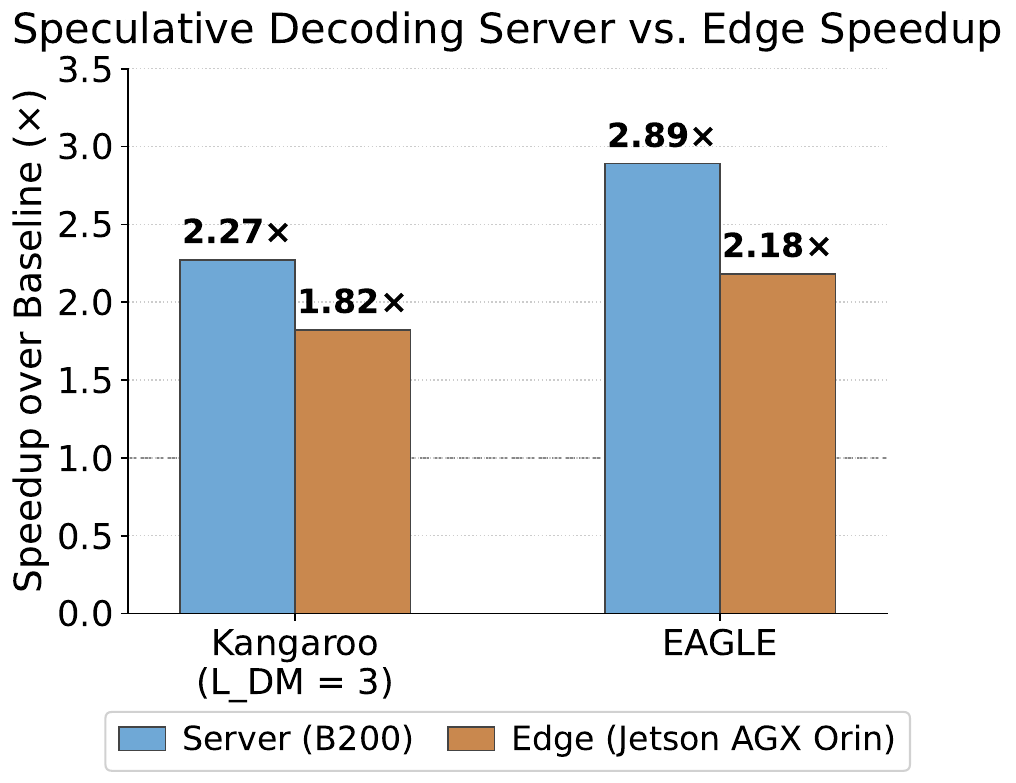}
     \vspace{-4mm}
\caption{End-to-end speedup of speculative decoding methods on B200 vs.\ Jetson AGX Orin.}
\label{fig:speed_compare_motivation}
\vspace{-2mm}
\end{wrapfigure}

shifts to this flash$\leftrightarrow$DRAM movement as reflected in the latency breakdown in Figure~\ref{fig:memory_hierarchy}. This bottleneck shift makes the existing speculative decoding methods inferior in memory-limited settings. We perform profiling on AGX Orin devices with limited memory budget and visualize this effect in Figure~\ref{fig:speed_compare_motivation}: the end-to-end speedup of Kangaroo and EAGLE drops by $19.8\%$ and $24.5\%$, respectively, when moving from server to edge, compared to vanilla inference on corresponding platforms. This effect is more severe for methods like EAGLE that introduce an additional draft model, since these components must also be staged through the constrained edge memory hierarchy.

\FloatBarrier
\section{Methodology}
\label{methodology}

In this section, we formalize the memory-adaptive three-stage inference framework that is central to \methodname{} and describe its three-stage inference pipeline. We then detail the verification tree construction and the Reduced KL Loss used to train the draft and shallow-verifier adapters. The detailed algorithm can be found in Appendix~\ref{app:algorithm}.

\subsection{Notation}

We use \emph{target model} (TM) for the target LLM, \emph{draft model} (DM) for layers $1$ to $L_\mathrm{DM}$ plus a lightweight adapter, which is well-trained to transform the shallow-layer outputs of the TM into calibrated representations suitable for next-token prediction, and \emph{shallow verifier} (SV) for layers $L_\mathrm{DM}{+}1$ to $L_\mathrm{SV}$ plus its adapter. A forward through layers $1$ to $L_\mathrm{DM}$ is a \emph{drafting pass}, through $L_\mathrm{DM}{+}1$ to $L_\mathrm{SV}$ an \emph{SV pass}, and through $L_\mathrm{SV}{+}1$ to $L_\mathrm{final}$ the \emph{final pass}; $\bar{\gamma}$ denotes the number of drafting passes per decoding cycle. Here, $L_\mathrm{DM}$ denotes the draft boundary, i.e., the last layer included in the draft model; $L_\mathrm{SV}$ denotes the shallow-verification boundary, i.e., the last layer used by the shallow verifier; and $L_\mathrm{final}$ denotes the final layer of the target model.

\subsection{Cascaded verification design}

\begin{figure}[!htbp]
    \centering
    \includegraphics[width=0.8\linewidth]{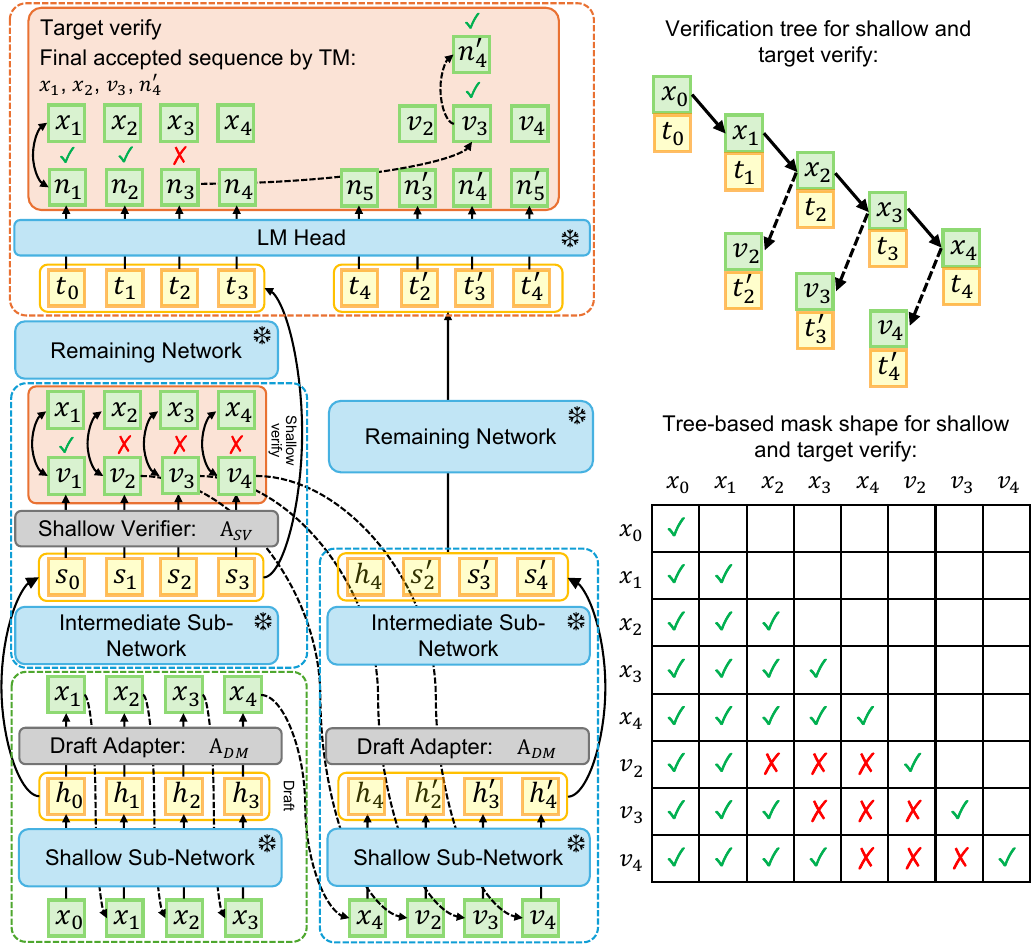}
    \caption{Pipeline of \methodname{}. \textbf{Left}: three-stage decoding cycle showing the interleaved flash$\leftrightarrow$DRAM transfers and GPU computation. The bottommost \textcolor{blue}{blue} block (draft sub-network) and the intermediate SV layers (middle \textcolor{blue}{blue} block) are streamed from flash once per full inference cycle during the drafting and shallow verification process (regarded as a SD process); the target layers (top \textcolor{blue}{blue} block, rest of the models) are streamed accordingly during the final pass. \textcolor{green}{Green} blocks: token embeddings; \textcolor{yellow}{yellow} blocks: hidden states; yellow-bordered box: parallel forward computation; \textcolor{orange}{orange} blocks: verification decision. Draft and SV adapter structure follows \citep{liu2024kangaroo}. \textbf{Right upper}: verification tree structure. \textbf{Right lower}: tree-based attention mask encoding the main draft sequence and SV correction branches for the batched final pass.}
    \label{fig:main_figure}
\end{figure}

The left part of Figure~\ref{fig:main_figure} illustrates the full \methodname{} pipeline. Each decoding cycle alternates between flash-to-DRAM weight movement and computation across three stages. In this paper, we focus on weight movement because edge inference is typically latency-oriented and operates at small batch sizes, where the KV cache is not the dominant memory consumer as in large-batch serving. When KV-cache pressure becomes relevant, it can be treated as part of the streamed state alongside model weights, so the same memory-adaptive scheduling principle still applies.

\textbf{Drafting loop.}
\methodname{} uses a compact DM that is small enough to fit within the DRAM budget of the edge device. The draft boundary $L_\mathrm{DM}$ is chosen not simply to maximize the number of layers that can be loaded into DRAM, but to balance three constraints: DRAM capacity, limited edge compute throughput, and memory bandwidth. A deeper draft model may improve the drafting quality, but it also increases the cost of each auto-regressive drafting step. We therefore keep the draft model intentionally shallow and perform $\bar{\gamma}$ sequential drafting passes to generate $\bar{\gamma}$ candidate tokens with reasonable latency. To preserve draft quality under this small-model constraint, the draft adapter is trained with the reduced KL loss introduced later.

\textbf{Shallow verification pass.}
The shallow verifier boundary is adaptively chosen according to memory budgets, which means that the medium sub-network (layers $L_\mathrm{DM}+1$ through $L_\mathrm{SV}$) will be moved to DRAM along with the draft model during the first flash to DRAM movement for each cycle. The $\bar{\gamma}$ candidates are then forwarded in parallel through these intermediate layers, and the SV adapter decodes a verification token at each position. Each SV token is compared strictly against the corresponding draft token: matching positions are passed through to the target stage; mismatching positions yield a correction candidate. When mismatches happen, the draft tokens and corrections are assembled into a verification tree as shown in Figure~\ref{fig:main_figure}.  The correction tokens are then re-forwarded from layer $1$ through $L_\mathrm{SV}$ under tree-masked attention to get the hidden states of the correction token---reusing the already-loaded draft and SV layers---at zero additional flash transfer cost. The insight of this DM-to-SV stage is to serve as an internal speculative decoding process that fully exploits the limited DRAM capacity.

\textbf{Target verification pass.} The remaining layers $L_\mathrm{SV}{+}1$ through $L_\mathrm{total}$ are then streamed from flash into DRAM chunk by chunk in the rest of the sequential transfer. The assembled verification tree is processed in a single batched forward pass under the corresponding tree attention mask. The target model can verify the main draft branch following greedy decoding or via typical acceptance~\cite{cai2024medusa}; at each rejected position, the SV's correction candidate is evaluated under the same criterion. The longest accepted prefix is committed to the output, and the KV cache is updated accordingly.

\subsection{Adapter design and training}

Draft and verification adapters follow the architecture of~\cite{liu2024kangaroo}: two normalization layers, one multi-head attention layer, and a shared LM head. Because \methodname{} uses compact draft and shallow-verifier sub-networks to meet edge device memory and latency constraints, their drafting and verification ability must be strengthened under limited sub-model capacity. Rather than distilling the full vocabulary from the full model~\cite{zhou2024distillspec,liu2024kangaroo}, which wastes supervision on low-probability tokens that rarely affect acceptance, we train the adapters with a \textbf{Reduced KL Loss} that focuses on the tokens most relevant to accepted tokens.

A general distillation loss that aligns the adapter distribution with the target model over the entire vocabulary is formed as:
{\small
\begin{equation}
    \mathcal{L}_{\text{full}} = -\frac{1}{|M|} \sum_{t \in M} \sum_{v=1}^{V}
    p_{\text{target}}(v \mid t) \cdot \log q_{\text{draft}}(v \mid t),
    \label{eq:full_loss}
\end{equation}
}
where $M$ is the set of unmasked token positions, $p_{\text{target}}$ is the target model's distribution, and $q_{\text{draft}}$ is the adapter's distribution. Spreading supervision across the full vocabulary $V$ dilutes the gradient signal onto near-zero-mass tokens that are irrelevant to accepted tokens. We instead restrict it to the top-$K$ tokens:
{\small
\begin{equation}
    \mathcal{L}_{\text{top-}K} = -\frac{1}{|M|} \sum_{t \in M} \sum_{v \in \mathcal{T}_K(t)}
    \tilde{p}_{\text{target}}(v \mid t) \cdot \log q_{\text{draft}}(v \mid t)
    \label{eq:topk_loss}
\end{equation}}
where $\mathcal{T}_K(t)$ denotes the set of top-$K$ tokens under the target distribution at position $t$, and $\tilde{p}_{\text{target}}(v \mid t)$ is the target distribution renormalized over this set. This concentrates distillation on the tokens that actually determine acceptance while excluding the low-probability tail. The detailed computation process can be found in Appendix~\ref{app:loss}.

\section{Experiments}
\label{experiments}

\textbf{Models, tasks, baselines, and metrics.} We conducted experiments on Vicuna models (7B and 13B)~\citep{chiang2023vicuna} and LLaMA2 models (7B and 13B)~\citep{touvron2023llama}, which are mainstream LLMs widely tested on different self-speculative decoding algorithms. We evaluated \methodname{} across multiple tasks, including multi-turn dialogue, code generation, mathematical reasoning, and instruction following, using Spec-bench~\citep{xia2024survey}, MT-bench~\citep{zheng2023judging}, GSM8K~\citep{cobbe2021training}, Alpaca~\citep{taori2023alpaca}, and HumanEval~\citep{chen2021evaluating}.

Unless otherwise stated, we set the drafting steps $\bar{\gamma}=5$, choose a compact draft boundary $L_\mathrm{DM}{=}3$, and set the shallow-verifier boundary to $L_\mathrm{SV}{=}15$. Following the design principle in Section~\ref{methodology}, $L_\mathrm{DM}$ is kept shallow to balance DRAM capacity, compute throughput, and repeated drafting cost, while $L_\mathrm{SV}{=}15$ corresponds to representative edge-memory budgets of 8\,GB for 7B models and 12\,GB for 13B models. We compare against representative speculative decoding algorithms, including REST~\citep{he2024rest}, Lookahead decoding~\citep{fu2024lookahead}, Kangaroo~\citep{liu2024kangaroo}, Medusa~\citep{cai2024medusa}, and EAGLE~\citep{li2024eagle}. Since our cascaded verification framework is orthogonal to EAGLE's high-probability candidate branching, \methodname{} w/ EAGLE adds these branches to our verification tree to enlarge the parallel candidate set without additional target-model forward passes.

We report the mean accepted length and end-to-end speedup. Mean accepted length measures the average number of tokens accepted per target-model verification, which is irrelevant to the deployed platform, while end-to-end speedup captures the wall-clock acceleration on the target platform. We use a server with 8 B200 GPUs to obtain mean accepted length results, and an NVIDIA Jetson AGX Orin as the edge device for end-to-end speedup evaluation.

\textbf{Training.} For fine-tuning our own adapters, we used the ShareGPT dataset~\citep{chiang2023vicuna} with 68,000 samples following Medusa, with learning rate set at 1e-6 for Vicuna 7B and 13B models, and 1e-5 for LLaMA2 7B and 13B models seperately. All the trainings are finished with 2 B200 GPUs within 13 hours.

\begin{table}[!htbp]
  \caption{Mean accepted length ($\tau$) and end-to-end speedup ($S\uparrow$) on Vicuna-7B across five benchmarks under greedy and relaxed (temperature\,=\,0.7) decoding.
    Kangaroo and \methodname{} use drafting steps\,=\,5; Medusa uses 2 heads; EAGLE is re-implemented in a self-speculative structure with drafting steps\,=\,5. In tree base decoding, we set top-$K$\,=\,10. Lookahead is greedy-only.}
  \vspace{-0.4em}
  \label{tab:speedup_temp_model}
  \centering
  \scriptsize
  \resizebox{\linewidth}{!}{%
  \begin{tabular}{ll cc cc cc cc cc cc}
    \toprule
    \textbf{Model} & \textbf{Algorithm}
      & \multicolumn{2}{c}{\textbf{Spec-bench}} & \multicolumn{2}{c}{\textbf{MT-bench}}
      & \multicolumn{2}{c}{\textbf{GSM8K}}      & \multicolumn{2}{c}{\textbf{Alpaca}}
      & \multicolumn{2}{c}{\textbf{HumanEval}}  & \multicolumn{2}{c}{\textbf{Mean}} \\
    \cmidrule(lr){3-4} \cmidrule(lr){5-6} \cmidrule(lr){7-8} \cmidrule(lr){9-10} \cmidrule(lr){11-12} \cmidrule(lr){13-14}
    & & $\tau$ & $S\uparrow$ & $\tau$ & $S\uparrow$ & $\tau$ & $S\uparrow$ & $\tau$ & $S\uparrow$ & $\tau$ & $S\uparrow$ & $\tau$ & $S\uparrow$ \\
    \midrule
    \multicolumn{14}{c}{Greedy} \\
    \midrule
    \raisebox{-8pt}[0pt][0pt]{\shortstack[l]{Vicuna-7b\\chain base}}
      & REST\citep{he2024rest}           & 1.6243 & 1.66$\times$ & 1.8239 & 1.86$\times$ & 1.5045 & 1.54$\times$ & 1.8406 & 1.88$\times$ & 1.7670 & 1.81$\times$ & 1.7121 & 1.75$\times$ \\
      & Lookahead\citep{fu2024lookahead} & 2.2785 & 2.29$\times$ & 2.2928 & 2.30$\times$ & 2.7010 & 2.72$\times$ & 2.0623 & 2.08$\times$ & 2.5128 & 2.53$\times$ & 2.3695 & 2.38$\times$ \\
      & Kangaroo\citep{liu2024kangaroo}  & 2.2736 & 1.75$\times$ & 2.4863 & 1.92$\times$ & 2.4856 & 1.92$\times$ & 2.3131 & 1.79$\times$ & 2.8979 & 2.24$\times$ & 2.4914 & 1.92$\times$ \\
      & Medusa\citep{cai2024medusa}      & 2.3116 & 2.32$\times$ & 2.5115 & 2.52$\times$ & 2.4966 & 2.51$\times$ & 2.3917 & 2.41$\times$ & 2.6832 & 2.70$\times$ & 2.4789 & 2.49$\times$ \\
      & EAGLE-self\citep{li2024eagle}   & 2.2995 & 1.74$\times$ & 2.5151 & 1.90$\times$ & 2.3213 & 1.76$\times$ & 2.3104 & 1.75$\times$ & 2.9005 & 2.19$\times$ & 2.4694 & 1.87$\times$ \\
      & \methodname{}                   & \textbf{2.8816} & \textbf{2.99$\times$} & \textbf{3.0520} & \textbf{3.16$\times$} & \textbf{3.0198} & \textbf{3.14$\times$} & \textbf{2.8280} & \textbf{2.94$\times$} & \textbf{3.5174} & \textbf{3.65$\times$} & \textbf{3.0598} & \textbf{3.18$\times$} \\
    \cmidrule(lr){1-14}
    \raisebox{-8pt}[0pt][0pt]{\shortstack[l]{Vicuna-7b\\tree base}}
      & EAGLE-self, tree                & 2.8943 & 2.18$\times$ & 3.1535 & 2.37$\times$ & 3.0059 & 2.26$\times$ & 2.9306 & 2.21$\times$ & 3.5949 & 2.70$\times$ & 3.1158 & 2.34$\times$ \\
      & \methodname{} w/ EAGLE          & \textbf{3.5097} & \textbf{3.51$\times$} & \textbf{3.6938} & \textbf{3.69$\times$} & \textbf{3.6869} & \textbf{3.70$\times$} & \textbf{3.4339} & \textbf{3.45$\times$} & \textbf{4.2009} & \textbf{4.21$\times$} & \textbf{3.7050} & \textbf{3.71$\times$} \\
    \midrule
    \multicolumn{14}{c}{Relaxed (temperature\,=\,0.7)} \\
    \midrule
    \raisebox{-8pt}[0pt][0pt]{\shortstack[l]{Vicuna-7b\\chain base}}
      & REST                            & 1.6236 & 1.66$\times$ & 1.8216 & 1.86$\times$ & 1.5192 & 1.55$\times$ & 1.8594 & 1.90$\times$ & 1.7984 & 1.84$\times$ & 1.7244 & 1.76$\times$ \\
      & Kangaroo                        & 2.7777 & 2.13$\times$ & 3.0712 & 2.35$\times$ & 3.0513 & 2.34$\times$ & 2.8348 & 2.18$\times$ & 3.3703 & 2.59$\times$ & 3.0211 & 2.32$\times$ \\
      & Medusa                          & 2.4884 & 2.50$\times$ & 2.6802 & 2.69$\times$ & 2.6837 & 2.70$\times$ & 2.6179 & 2.63$\times$ & 2.8760 & 2.89$\times$ & 2.6692 & 2.68$\times$ \\
      & EAGLE-self                      & 2.7487 & 2.08$\times$ & 3.0753 & 2.33$\times$ & 2.9515 & 2.23$\times$ & 2.8644 & 2.17$\times$ & 3.4546 & 2.61$\times$ & 3.0189 & 2.28$\times$ \\
      & \methodname{}                   & \textbf{3.4791} & \textbf{3.60$\times$} & \textbf{3.7597} & \textbf{3.89$\times$} & \textbf{3.9652} & \textbf{4.12$\times$} & \textbf{3.6742} & \textbf{3.82$\times$} & \textbf{4.0802} & \textbf{4.24$\times$} & \textbf{3.7917} & \textbf{3.94$\times$} \\
    \cmidrule(lr){1-14}
    \raisebox{-8pt}[0pt][0pt]{\shortstack[l]{Vicuna-7b\\tree base}}
      & EAGLE-self                      & 3.3494 & 2.53$\times$ & 3.7271 & 2.82$\times$ & 3.7629 & 2.85$\times$ & 3.5334 & 2.68$\times$ & 4.1547 & 3.14$\times$ & 3.7055 & 2.80$\times$ \\
      & \methodname{} w/ EAGLE          & \textbf{4.1802} & \textbf{4.18$\times$} & \textbf{4.5229} & \textbf{4.52$\times$} & \textbf{4.5345} & \textbf{4.55$\times$} & \textbf{4.2745} & \textbf{4.29$\times$} & \textbf{4.8408} & \textbf{4.85$\times$} & \textbf{4.4706} & \textbf{4.48$\times$} \\
    \bottomrule
  \end{tabular}%
  }
\end{table}

\subsection{Main results}

\paragraph{Mean accepted length and end-to-end speedup.}
Tables~\ref{tab:speedup_temp_model} and~\ref{tab:speedup_other_models} report the main results. Since EAGLE was originally designed around feature-level speculative decoding, we re-implement it in our self-speculative setting and evaluate both decoding modes: chain decoding and tree decoding. In the edge setting, where each decoding loop is constrained by limited compute and weight-transfer bandwidth, increasing the mean accepted length directly reduces the number of costly inference loops and is reflected in the measured end-to-end speedup.

\begin{table}[!htbp]
  \caption{Mean accepted length ($\tau$) and end-to-end speedup ($S\uparrow$) on Vicuna-13B and LLaMA2-7B/13B under greedy and relaxed (temperature\,=\,0.7) decoding. Methods and settings follow Table~\ref{tab:speedup_temp_model}.}
  \vspace{-0.4em}
  \label{tab:speedup_other_models}
  \centering
  \scriptsize
  \resizebox{\linewidth}{!}{%
  \begin{tabular}{ll cc cc cc cc cc cc}
    \toprule
    \textbf{Model} & \textbf{Algorithm}
      & \multicolumn{2}{c}{\textbf{Spec-bench}} & \multicolumn{2}{c}{\textbf{MT-bench}}
      & \multicolumn{2}{c}{\textbf{GSM8K}}      & \multicolumn{2}{c}{\textbf{Alpaca}}
      & \multicolumn{2}{c}{\textbf{HumanEval}}  & \multicolumn{2}{c}{\textbf{Mean}} \\
    \cmidrule(lr){3-4} \cmidrule(lr){5-6} \cmidrule(lr){7-8} \cmidrule(lr){9-10} \cmidrule(lr){11-12} \cmidrule(lr){13-14}
    & & $\tau$ & $S\uparrow$ & $\tau$ & $S\uparrow$ & $\tau$ & $S\uparrow$ & $\tau$ & $S\uparrow$ & $\tau$ & $S\uparrow$ & $\tau$ & $S\uparrow$ \\
    \midrule
    \multicolumn{14}{c}{Greedy} \\
    \midrule
    \raisebox{-8pt}[0pt][0pt]{\shortstack[l]{Vicuna-13b\\chain base}}
      & Kangaroo   & 2.1937 & 2.19$\times$ & 2.4184 & 2.41$\times$ & 2.3848 & 2.38$\times$ & 2.2226 & 2.22$\times$ & 2.9183 & 2.91$\times$ & 2.4276 & 2.42$\times$ \\
      & Medusa     & 2.3920 & 2.03$\times$ & 2.5833 & 2.19$\times$ & 2.5824 & 2.20$\times$ & 2.3959 & 2.04$\times$ & 2.8035 & 2.39$\times$ & 2.5514 & 2.17$\times$ \\
      & EAGLE-self & 2.1952 & 2.18$\times$ & 2.3831 & 2.37$\times$ & 2.2031 & 2.19$\times$ & 2.2183 & 2.21$\times$ & 2.9163 & 2.90$\times$ & 2.3832 & 2.37$\times$ \\
      & \methodname{} & \textbf{2.6588} & \textbf{2.64$\times$} & \textbf{2.7946} & \textbf{2.77$\times$} & \textbf{2.7636} & \textbf{2.74$\times$} & \textbf{2.5493} & \textbf{2.53$\times$} & \textbf{3.2218} & \textbf{3.20$\times$} & \textbf{2.7976} & \textbf{2.78$\times$} \\
    \cmidrule(lr){1-14}
    \raisebox{-8pt}[0pt][0pt]{\shortstack[l]{Vicuna-13b\\tree base}}
      & EAGLE-self, tree  & 2.8035 & 2.77$\times$ & 3.0423 & 3.01$\times$ & 2.8760 & 2.85$\times$ & 2.8337 & 2.81$\times$ & 3.6168 & 3.58$\times$ & 3.0345 & 3.00$\times$ \\
      & \methodname{} w/ EAGLE & \textbf{3.4708} & \textbf{3.42$\times$} & \textbf{3.6897} & \textbf{3.64$\times$} & \textbf{3.5481} & \textbf{3.50$\times$} & \textbf{3.3568} & \textbf{3.32$\times$} & \textbf{4.2428} & \textbf{4.19$\times$} & \textbf{3.6616} & \textbf{3.61$\times$} \\
    \cmidrule(lr){1-14}
    \raisebox{-8pt}[0pt][0pt]{\shortstack[l]{Llama2-7b\\chain base}}
      & Kangaroo   & 3.8176 & 3.29$\times$ & 4.1930 & 3.61$\times$ & 4.4296 & 3.82$\times$ & 4.0431 & 3.49$\times$ & 4.2946 & 3.70$\times$ & 4.1534 & 3.58$\times$ \\
      & Medusa     & 1.6660 & 1.67$\times$ & 1.7817 & 1.78$\times$ & 1.7358 & 1.74$\times$ & 1.7758 & 1.78$\times$ & 1.9657 & 1.97$\times$ & 1.7850 & 1.78$\times$ \\
      & EAGLE-self & 3.8068 & 3.28$\times$ & 4.1813 & 3.60$\times$ & 4.7597 & 4.10$\times$ & 4.0429 & 3.49$\times$ & 4.0588 & 3.50$\times$ & 4.1699 & 3.60$\times$ \\
      & \methodname{} & \textbf{4.3335} & \textbf{4.33$\times$} & \textbf{4.6181} & \textbf{4.62$\times$} & \textbf{5.1055} & \textbf{5.10$\times$} & \textbf{4.4642} & \textbf{4.46$\times$} & \textbf{4.7240} & \textbf{4.72$\times$} & \textbf{4.6491} & \textbf{4.65$\times$} \\
    \cmidrule(lr){1-14}
    \raisebox{-8pt}[0pt][0pt]{\shortstack[l]{Llama2-7b\\tree base}}
      & EAGLE-self, tree  & 4.3529 & 3.75$\times$ & 4.7199 & 4.07$\times$ & 5.0954 & 4.39$\times$ & 4.5393 & 3.91$\times$ & 4.5866 & 3.95$\times$ & 4.6588 & 4.02$\times$ \\
      & \methodname{} w/ EAGLE & \textbf{4.8318} & \textbf{4.83$\times$} & \textbf{5.0634} & \textbf{5.06$\times$} & \textbf{5.4185} & \textbf{5.42$\times$} & \textbf{4.9101} & \textbf{4.91$\times$} & \textbf{5.1741} & \textbf{5.17$\times$} & \textbf{5.0796} & \textbf{5.08$\times$} \\
    \cmidrule(lr){1-14}
    \raisebox{-8pt}[0pt][0pt]{\shortstack[l]{Llama2-13b\\chain base}}
      & Kangaroo   & 3.5409 & 3.53$\times$ & 3.9369 & 3.92$\times$ & 4.2421 & 4.23$\times$ & 3.9754 & 3.97$\times$ & 4.1133 & 4.10$\times$ & 3.9617 & 3.95$\times$ \\
      & Medusa     & 1.8250 & 1.55$\times$ & 1.9737 & 1.68$\times$ & 1.9721 & 1.68$\times$ & 1.9425 & 1.65$\times$ & 1.9425 & 1.65$\times$ & 1.9312 & 1.64$\times$ \\
      & EAGLE-self & 3.5478 & 3.53$\times$ & 3.9754 & 3.96$\times$ & 3.8572 & 3.84$\times$ & 3.9805 & 3.97$\times$ & 4.1082 & 4.09$\times$ & 3.8938 & 3.88$\times$ \\
      & \methodname{} & \textbf{4.0570} & \textbf{4.02$\times$} & \textbf{4.2752} & \textbf{4.24$\times$} & \textbf{4.3415} & \textbf{4.31$\times$} & \textbf{4.3696} & \textbf{4.34$\times$} & \textbf{4.6959} & \textbf{4.66$\times$} & \textbf{4.3478} & \textbf{4.32$\times$} \\
    \cmidrule(lr){1-14}
    \raisebox{-8pt}[0pt][0pt]{\shortstack[l]{Llama2-13b\\tree base}}
      & EAGLE-self, tree  & 4.0829 & 4.04$\times$ & 4.4699 & 4.42$\times$ & 4.2793 & 4.24$\times$ & 4.4566 & 4.42$\times$ & 4.6993 & 4.65$\times$ & 4.3976 & 4.35$\times$ \\
      & \methodname{} w/ EAGLE & \textbf{4.5707} & \textbf{4.51$\times$} & \textbf{4.8191} & \textbf{4.75$\times$} & \textbf{4.8534} & \textbf{4.79$\times$} & \textbf{4.7998} & \textbf{4.75$\times$} & \textbf{5.0178} & \textbf{4.95$\times$} & \textbf{4.8122} & \textbf{4.75$\times$} \\
    \midrule
    \multicolumn{14}{c}{Relaxed (temperature\,=\,0.7)} \\
    \midrule
    \raisebox{-8pt}[0pt][0pt]{\shortstack[l]{Vicuna-13b\\chain base}}
      & Kangaroo   & 2.4889 & 2.48$\times$ & 2.7894 & 2.78$\times$ & 2.7695 & 2.76$\times$ & 2.5927 & 2.59$\times$ & 3.2753 & 3.27$\times$ & 2.7832 & 2.78$\times$ \\
      & Medusa     & 2.5852 & 2.20$\times$ & 2.7905 & 2.37$\times$ & 2.7920 & 2.38$\times$ & 2.7148 & 2.31$\times$ & 2.9535 & 2.51$\times$ & 2.7672 & 2.35$\times$ \\
      & EAGLE-self & 2.5144 & 2.50$\times$ & 2.8444 & 2.83$\times$ & 2.5893 & 2.58$\times$ & 2.6609 & 2.65$\times$ & 3.2212 & 3.20$\times$ & 2.7660 & 2.75$\times$ \\
      & \methodname{} & \textbf{2.9743} & \textbf{2.95$\times$} & \textbf{3.2421} & \textbf{3.22$\times$} & \textbf{3.0736} & \textbf{3.05$\times$} & \textbf{2.9963} & \textbf{2.98$\times$} & \textbf{3.4792} & \textbf{3.45$\times$} & \textbf{3.1531} & \textbf{3.13$\times$} \\
    \cmidrule(lr){1-14}
    \raisebox{-8pt}[0pt][0pt]{\shortstack[l]{Vicuna-13b\\tree base}}
      & EAGLE-self, tree  & 3.1934 & 3.16$\times$ & 3.5232 & 3.49$\times$ & 3.3631 & 3.33$\times$ & 3.3051 & 3.28$\times$ & 3.9784 & 3.94$\times$ & 3.4726 & 3.44$\times$ \\
      & \methodname{} w/ EAGLE & \textbf{3.9851} & \textbf{3.93$\times$} & \textbf{4.2360} & \textbf{4.18$\times$} & \textbf{4.1399} & \textbf{4.09$\times$} & \textbf{4.0683} & \textbf{4.02$\times$} & \textbf{4.7269} & \textbf{4.67$\times$} & \textbf{4.2312} & \textbf{4.18$\times$} \\
    \cmidrule(lr){1-14}
    \raisebox{-8pt}[0pt][0pt]{\shortstack[l]{Llama2-7b\\chain base}}
      & Kangaroo   & 4.4849 & 3.87$\times$ & 4.9609 & 4.28$\times$ & 4.8506 & 4.18$\times$ & 4.9796 & 4.29$\times$ & 4.0587 & 3.50$\times$ & 4.6669 & 4.02$\times$ \\
      & Medusa     & 1.7482 & 1.75$\times$ & 1.8667 & 1.87$\times$ & 2.1875 & 2.19$\times$ & 1.8955 & 1.90$\times$ & 2.2848 & 2.28$\times$ & 1.9965 & 2.00$\times$ \\
      & EAGLE-self & 4.4910 & 3.87$\times$ & 4.9756 & 4.29$\times$ & 4.9139 & 4.24$\times$ & 4.9867 & 4.30$\times$ & 4.2687 & 3.68$\times$ & 4.7272 & 4.08$\times$ \\
      & \methodname{} & \textbf{4.8831} & \textbf{4.88$\times$} & \textbf{5.2004} & \textbf{5.20$\times$} & \textbf{5.1870} & \textbf{5.19$\times$} & \textbf{5.1976} & \textbf{5.20$\times$} & \textbf{4.4388} & \textbf{4.44$\times$} & \textbf{4.9814} & \textbf{4.98$\times$} \\
    \cmidrule(lr){1-14}
    \raisebox{-8pt}[0pt][0pt]{\shortstack[l]{Llama2-7b\\tree base}}
      & EAGLE-self, tree  & 4.9713 & 4.29$\times$ & 5.2409 & 4.52$\times$ & 5.2814 & 4.55$\times$ & 5.3228 & 4.59$\times$ & 4.7703 & 4.11$\times$ & 5.1173 & 4.41$\times$ \\
      & \methodname{} w/ EAGLE & \textbf{5.3143} & \textbf{5.31$\times$} & \textbf{5.5818} & \textbf{5.58$\times$} & \textbf{5.5096} & \textbf{5.51$\times$} & \textbf{5.5601} & \textbf{5.56$\times$} & \textbf{4.9882} & \textbf{4.99$\times$} & \textbf{5.3908} & \textbf{5.39$\times$} \\
    \cmidrule(lr){1-14}
    \raisebox{-8pt}[0pt][0pt]{\shortstack[l]{Llama2-13b\\chain base}}
      & Kangaroo   & 4.1486 & 4.14$\times$ & 4.6991 & 4.68$\times$ & 4.6031 & 4.59$\times$ & 4.7376 & 4.73$\times$ & 4.6950 & 4.68$\times$ & 4.5767 & 4.56$\times$ \\
      & Medusa     & 1.9322 & 1.64$\times$ & 2.0646 & 1.75$\times$ & 2.2499 & 1.92$\times$ & 2.0718 & 1.76$\times$ & 2.2848 & 1.95$\times$ & 2.1207 & 1.80$\times$ \\
      & EAGLE-self & 4.1452 & 4.12$\times$ & 4.7032 & 4.68$\times$ & 4.5883 & 4.57$\times$ & 4.7401 & 4.72$\times$ & 4.7135 & 4.69$\times$ & 4.5781 & 4.56$\times$ \\
      & \methodname{} & \textbf{4.6461} & \textbf{4.61$\times$} & \textbf{4.9133} & \textbf{4.87$\times$} & \textbf{5.3102} & \textbf{5.27$\times$} & \textbf{4.9464} & \textbf{4.92$\times$} & \textbf{5.0678} & \textbf{5.03$\times$} & \textbf{4.9768} & \textbf{4.94$\times$} \\
    \cmidrule(lr){1-14}
    \raisebox{-8pt}[0pt][0pt]{\shortstack[l]{Llama2-13b\\tree base}}
      & EAGLE-self, tree  & 4.7176 & 4.67$\times$ & 5.1051 & 5.05$\times$ & 5.1449 & 5.10$\times$ & 5.0338 & 4.99$\times$ & 5.1758 & 5.12$\times$ & 5.0354 & 4.98$\times$ \\
      & \methodname{} w/ EAGLE & \textbf{5.0813} & \textbf{5.01$\times$} & \textbf{5.3499} & \textbf{5.28$\times$} & \textbf{5.5524} & \textbf{5.48$\times$} & \textbf{5.3241} & \textbf{5.27$\times$} & \textbf{5.3969} & \textbf{5.33$\times$} & \textbf{5.3409} & \textbf{5.27$\times$} \\
    \bottomrule
  \end{tabular}%
  }
\end{table}

Table~\ref{tab:speedup_temp_model} reports results on Vicuna-7B under both greedy and relaxed decoding. Under greedy decoding, \methodname{} accepts $3.06$ tokens per target call on average, compared with $2.49$ for Kangaroo, $2.48$ for Medusa, and $2.47$ for our self-speculative EAGLE chain baseline---a $23$--$24\%$ relative gain. This translates into a $3.18\times$ end-to-end speedup, $66\%$ higher than Kangaroo's $1.92\times$. Compared with the full-baseline Lookahead and REST, \methodname{} improves mean accepted length by $29\%$ and $79\%$, respectively. Under relaxed decoding ($\tau{=}0.7$), these gains are maintained or amplified: \methodname{} continues to lead all baselines including REST, whose relaxed-acceptance advantage over greedy is outpaced by the longer draft sequences that \methodname{} produces. Table~\ref{tab:speedup_other_models} extends the evaluation to Vicuna-13B and LLaMA2-7B/13B under both decoding modes. On LLaMA2-7B, \methodname{} reaches $4.65$ accepted tokens and $4.65\times$ speedup under greedy decoding, versus $4.17$ tokens ($3.60\times$) for the EAGLE chain baseline and $4.15$ tokens ($3.58\times$) for Kangaroo. On the larger LLaMA2-13B, \methodname{} achieves $4.35$ accepted tokens and $4.32\times$ speedup, compared with $3.89$ ($3.88\times$) for EAGLE-self and $3.96$ ($3.95\times$) for Kangaroo. Pairing \methodname{} with EAGLE tree decoding further pushes accepted length to $5.08$ and $4.81$ tokens on LLaMA2-7B and 13B, respectively, with corresponding speedups of $5.08\times$ and $4.75\times$. These consistent gains across model families and sizes confirm that the cascaded verification advantage is not model-specific.

\begin{figure}[!htbp]
    \centering
    \includegraphics[width=0.9\linewidth]{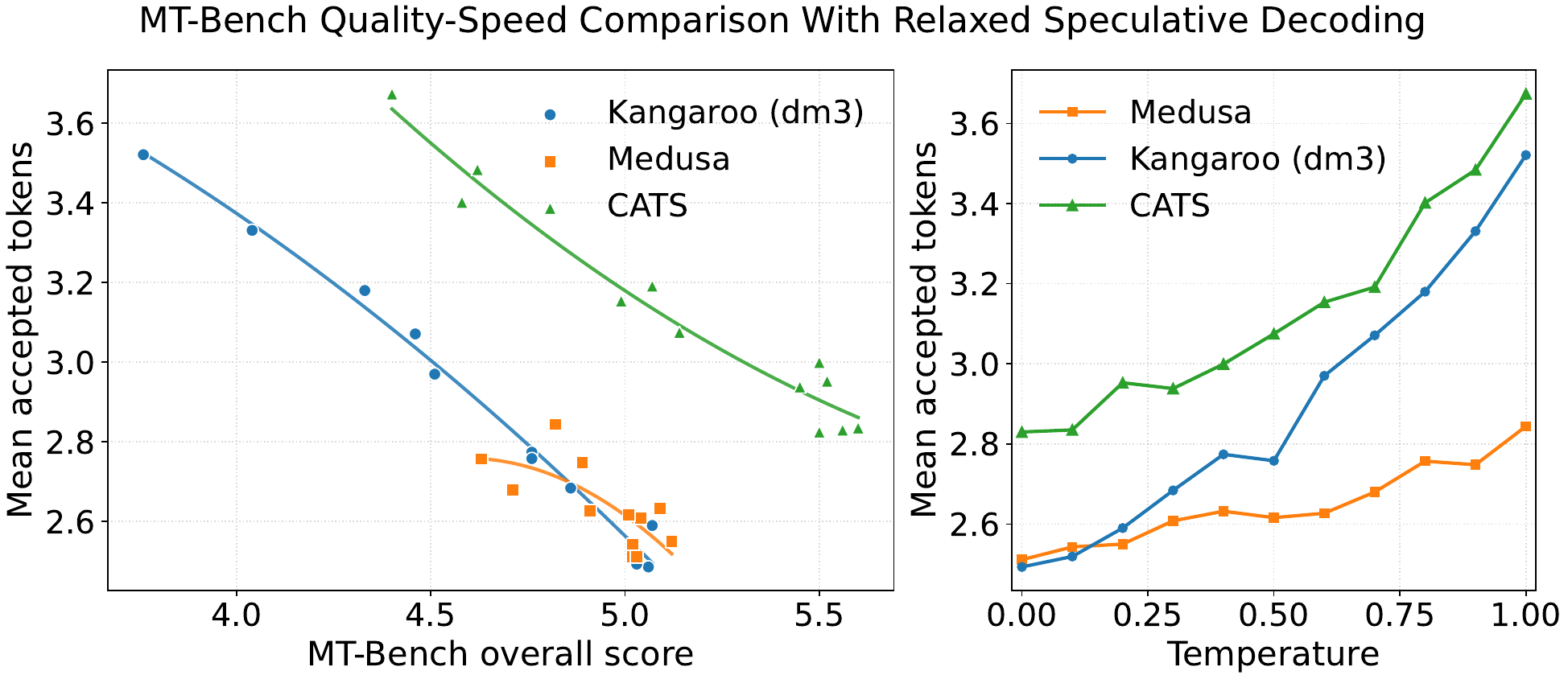}
    \vspace{-0.8em}
    \caption{MT-Bench quality–speed comparison under relaxed acceptance. Quality is measured by the GPT-4o following the protocol of \citep{zheng2023judging}; we report the mean score across all MT-Bench categories.}
    \label{fig:mt_bench_compare}
\end{figure}

\paragraph{Quality--speed Pareto frontier.}
A common concern with relaxed speculative decoding is that increasing the number of accepted draft tokens may improve speed at the expense of generation quality. Figure~\ref{fig:mt_bench_compare} shows that this trade-off is not inherent to \methodname{}. Under matched schedules ($\tau{=}0.3$, $\alpha{=}0.09$, $5$ draft steps), \methodname{} lies on the upper-right Pareto frontier on MT-Bench, improving accepted length while maintaining competitive evaluation scores. Sweeping temperature from $0$ to $1$ preserves the separation, suggesting that cascaded verification increases usable parallelism without relaxing the target distribution.

\FloatBarrier
\subsection{Ablation Study}
\label{sec:ablation}

In ablation studies, we examine how \methodname{} behaves under the two constraints that dominate limited memory deployment: memory budget and auto-regressive drafting cost. Unless otherwise stated, all ablations use Vicuna-7B evaluated on MT-Bench. We first analyze how to capture the wall-clock advantage of the memory-adaptive design, and then study how the drafting horizon $\bar{\gamma}$ trades off accepted length against additional loop latency and memory traffic.

\subsubsection{Memory Budget Analysis}
\label{sec:ablation_chunk}

\paragraph{Adaptivity across memory budgets.}
A key advantage of \methodname{} is that the shallow-verifier depth $L_\mathrm{SV}$ is a free parameter (Section~\ref{sec:ablation_chunk}, properties (i)--(ii)) and can therefore be tuned to fully consume whatever DRAM budget the deployment platform offers, without changing the rest of the pipeline. We exercise this property on three representative edge-memory budgets covering tightly to comfortably resourced devices---$2$\,GB, $6$\,GB, and $8$\,GB---and at each budget configure \methodname{} with the deepest $L_\mathrm{SV}$ that fits. Table~\ref{tab:memory_budget_ablation} reports per-token compute time, mean accepted length, tokens/s, and end-to-end speedup over the non-speculative baseline. Across all three regimes, \methodname{} achieves the highest speedup, demonstrating that its advantage is not an artifact of one specific memory configuration but a structural benefit that adapts to the available budget.

\begin{table}[!htbp]
    \caption{Acceleration under different memory budgets on Vicuna-7B.}
    \vspace{-0.4em}
    \label{tab:memory_budget_ablation}
    \centering
    \footnotesize
    \resizebox{0.82\linewidth}{!}{%
    \begin{tabular}{llcccc}
        \toprule
        \textbf{Budget} & \textbf{Method} & \textbf{Comp/tok (s)} & \textbf{Mean acc.} & \textbf{Tok/s$\uparrow$} & \textbf{Speedup} \\
        \midrule
        2\,GB & Baseline & 0.213 & 1.00 & 0.117 & $1.00\times$ \\
        2\,GB & Kangaroo ($L_\mathrm{DM}{=}3$) & 0.094 & 2.27 & 0.266 & $2.27\times$ \\
        2\,GB & \methodname{} ($L_\mathrm{DM}{=}3$, $L_\mathrm{SV}{=}5$) & \textbf{0.067} & \textbf{2.80} & \textbf{0.329} & $\mathbf{2.82\times}$ \\
        \midrule
        6\,GB & Baseline & 0.204 & 1.00 & 0.126 & $1.00\times$ \\
        6\,GB & Kangaroo ($L_\mathrm{DM}{=}3$) & 0.089 & 2.27 & 0.267 & $2.12\times$ \\
        6\,GB & \methodname{} ($L_\mathrm{DM}{=}3$, $L_\mathrm{SV}{=}10$) & \textbf{0.064} & \textbf{2.94} & \textbf{0.373} & $\mathbf{2.96\times}$ \\
        \midrule
        8\,GB & Baseline & 0.253 & 1.00 & 0.146 & $1.00\times$ \\
        8\,GB & Kangaroo ($L_\mathrm{DM}{=}3$) & 0.098 & 2.27 & 0.282 & $1.93\times$ \\
        8\,GB & \methodname{} ($L_\mathrm{DM}{=}3$, $L_\mathrm{SV}{=}15$) & \textbf{0.062} & \textbf{3.05} & \textbf{0.461} & $\mathbf{3.16\times}$ \\
        \bottomrule
    \end{tabular}}
\end{table}
\vspace{-0.5em}

A natural fix for Kangaroo's limited acceptance is to deepen its draft model. However, because the draft path is invoked $\bar{\gamma}$ times per cycle, deepening it amplifies both the repeated DRAM$\leftrightarrow$on-chip transfer cost and the per-cycle compute load---an overhead that is non-negligible on edge devices. We quantify this with \emph{Bytes Per Token} (BPT)~\citep{alizadeh2024llmflash}, the average weight volume streamed from DRAM per generated token. For a two-stage Kangaroo-style pipeline,
{\small
\begin{equation}
    \mathrm{BPT_{Kangaroo}} = \frac{\bar{\gamma} \cdot B_{\mathrm{draft}} + B_{\mathrm{verify}}}{\bar{\beta} + 1},
    \label{eq:bpt}
\end{equation}}
where $B_{\mathrm{draft}}$, $B_{\mathrm{verify}}$ are draft and verification parameter volumes and $\bar{\gamma}$ is drafting passes per cycle. \methodname{}'s three-stage pipeline inserts a shallow verifier, splitting $B_{\mathrm{verify}}$ into two contiguous segments:
{\small
\begin{equation}
    \mathrm{BPT_{CATS}} = \frac{\bar{\gamma} \cdot B_{\mathrm{draft}} + B_{\mathrm{SV}} + B_{\mathrm{target}}}{\bar{\beta} + 1},
    \label{eq:bpt_sv}
\end{equation}}
with $B_{\mathrm{SV}}+B_{\mathrm{target}}=B_{\mathrm{verify}}$. Because $B_{\mathrm{SV}}$ is streamed once per cycle as part of the target forward pass, enlarging $L_\mathrm{SV}$ increases verifier capacity without inflating the $\bar{\gamma}\cdot B_{\mathrm{draft}}$ term.

Table~\ref{tab:edge_orin} validates this on Jetson AGX Orin (Vicuna-7B, $8$\,GB budget). Deepening Kangaroo from $L_\mathrm{DM}{=}3$ to $15$ raises $\bar{\beta}{+}1$ from $2.27$ to $2.98$, yet speedup \emph{falls} from $1.93\times$ to $1.53\times$ as BPT climbs from $8.37$ to $11.10$\,GB/tok. \methodname{} ($L_\mathrm{DM}{=}3$, $L_\mathrm{SV}{=}15$) instead achieves similar acceptance ($3.05$) with the lowest BPT ($5.84$\,GB/tok), and the best speedup ($3.16\times$), confirming that intermediate-layer verification---not a deeper drafter---is the key to edge throughput.

\begin{table}[!htbp]
    \caption{Same-budget edge ablation on NVIDIA AGX Orin (Vicuna-7B). Deepening Kangaroo's draft model improves accepted length but increases BPT and compute cost, while \methodname{} achieves the best speedup with a shallow drafter and single-pass shallow verification.}
    \vspace{-0.4em}
    \label{tab:edge_orin}
    \centering
    \footnotesize
    \resizebox{0.85\linewidth}{!}{%
    \begin{tabular}{lccccc}
        \toprule
        \textbf{Method} & \textbf{BPT (GB/tok)$\downarrow$} & \textbf{Comp/tok (s)} & \textbf{Mean acc.} & \textbf{Tok/s$\uparrow$} & \textbf{Speedup} \\
        \midrule
        Baseline ($L{=}32$)                                         & 12.95          & 0.253          & 1.00          & 0.146          & $1.00\times$ \\
        Kangaroo ($L_\mathrm{DM}{=}3$)                             & 8.37          & 0.098          & 2.27          & 0.282           & $1.93\times$ \\
        Kangaroo ($L_\mathrm{DM}{=}5$)                             & 8.67          & 0.090          & 2.41          & 0.275          & $1.88\times$ \\
        Kangaroo ($L_\mathrm{DM}{=}15$)                            & 11.10          & 0.106          & 2.98          & 0.224          & $1.53\times$ \\
        \midrule
        \methodname{} ($L_\mathrm{DM}{=}3$, $L_\mathrm{SV}{=}15$) & \textbf{5.84} & \textbf{0.062} & \textbf{3.05} & \textbf{0.461} & $\mathbf{3.16\times}$ \\
        \bottomrule
    \end{tabular}}
\end{table}

\vspace{-0.5em}

\subsubsection{Drafting Steps Ablation}
\label{sec:ablation_gamma}

We next vary the drafting horizon $\bar{\gamma}$ to justify the default choice $\bar{\gamma}{=}5$. Increasing $\bar{\gamma}$ gives the draft stage more opportunities to propose tokens, but it also lengthens the autoregressive drafting loop and increases the streamed volume in the BPT numerator. Table~\ref{tab:ablation_gamma} shows this trade-off: moving \methodname{} from $\bar{\gamma}{=}3$ to $5$ substantially improves accepted length from $2.7077$ to $3.0520$ with only a small BPT increase, whereas further extending the horizon to $7$ or $10$ yields diminishing acceptance gains while BPT continues to rise. We therefore use $\bar{\gamma}{=}5$ as a balanced setting that captures most of the acceptance benefit before the drafting horizon reaches saturation, while avoiding unnecessary loop latency and memory traffic.

\begin{table}[!htbp]
    \caption{Drafting-horizon ablation on Vicuna-7B, reporting accepted length, BPT, and speedup.}
    \vspace{-0.4em}
    \label{tab:ablation_gamma}
    \centering
    \footnotesize
    \resizebox{0.85\linewidth}{!}{%
    \begin{tabular}{c ccc ccc ccc}
        \toprule
        & \multicolumn{3}{c}{Kangaroo} & \multicolumn{3}{c}{EAGLE-self} & \multicolumn{3}{c}{\methodname{}} \\
        \cmidrule(lr){2-4} \cmidrule(lr){5-7} \cmidrule(lr){8-10}
        $\bar{\gamma}$ & \textbf{Mean acc.} & \textbf{BPT$\downarrow$} & \textbf{Speedup} & \textbf{Mean acc.} & \textbf{BPT$\downarrow$} & \textbf{Speedup} & \textbf{Mean acc.} & \textbf{BPT$\downarrow$} & \textbf{Speedup} \\
        \midrule
        3  & 2.2501 & 11.1531 & $1.75\times$ & 2.2475 & 11.1660 & $1.73\times$ & \textbf{2.7077} & \textbf{5.6805} & $\mathbf{2.82\times}$ \\
        5  & 2.4863 & 13.3496 & $1.92\times$ & 2.5151 & 13.1967 & $1.90\times$ & \textbf{3.0520} & \textbf{5.8354} & $\mathbf{3.16\times}$ \\
        7  & 2.6330 & 15.6804 & $2.04\times$ & 2.6194 & 15.7618 & $2.00\times$ & \textbf{3.2343} & \textbf{6.2574} & $\mathbf{3.34\times}$ \\
        10 & 2.7142 & 19.6852 & $2.09\times$ & 2.6984 & 19.8004 & $2.05\times$ & \textbf{3.3332} & \textbf{7.1646} & $\mathbf{3.43\times}$ \\
        \bottomrule
    \end{tabular}}
\end{table}
\vspace{-0.5em}

\FloatBarrier
\section{Conclusion and Limitations}
\label{sec:conclusion}

\methodname{} addresses the flash-transfer bottleneck of memory-limited LLM inference through a staged verification cascade adapted to the device's DRAM capacity and weight-offloading schedule, raising both token acceptance and end-to-end speedup while keeping the on-device memory footprint equal to that of the target model. Across four target models and five benchmarks, \methodname{} delivers up to $5.08\times$ wall-clock speedup with no quality degradation. Several limitations remain: our evaluation is restricted to $7$B--$13$B models; the draft sub-network and SV adapter rely on distilled adapters, incurring an up-front training cost; and the three-stage design presumes transformer-style layer-aligned representations, restricting direct applicability to architectures such as state-space models.


{
\small

\bibliographystyle{unsrt}
\bibliography{references}

}

\clearpage

\appendix

\section{\methodname{} algorithm}\label{app:algorithm}

\begin{algorithm}[h]
\footnotesize
\caption{\methodname{}: Cascaded Self-Speculative Decoding with Tree-Masked Final Verification}
\label{alg:cats}
\begin{algorithmic}[1]
\Require Prompt $\mathbf{x}$; target model $\mathcal{M}$ with layers $1{:}L_\mathrm{total}$ and LM head $\mathcal{H}$;
draft cut-point $L_\mathrm{DM}$, shallow-verifier cut-point $L_\mathrm{SV}$
($1{\le}L_\mathrm{DM}{<}L_\mathrm{SV}{<}L_\mathrm{total}$);
draft adapter $\mathcal{A}_\mathrm{DM}$, SV adapter $\mathcal{A}_\mathrm{SV}$;
drafting horizon $\bar{\gamma}$; max new tokens $T$; acceptance criterion $\textsc{Acc}$ (greedy or typical~\cite{cai2024medusa})
\Ensure Generated sequence $\mathbf{y}$
\Statex
\State \textbf{// ===== Initialization =====}
\State Load layers $1{:}L_\mathrm{SV}$ (DM\,$\cup$\,SV) into DRAM and pin resident
\State $(\mathbf{h}^{(1{:}L_\mathrm{total})},\, \mathrm{KV}_\mathcal{M}) \gets \mathcal{M}(\mathbf{x})$ \Comment{prompt prefill, full target}
\State $y_0 \gets \arg\max \mathcal{H}(\mathbf{h}^{(L_\mathrm{total})}_{-1})$;\quad
       initialize $\mathrm{KV}_{\mathcal{A}_\mathrm{DM}},\, \mathrm{KV}_{\mathcal{A}_\mathrm{SV}}$ from $\mathbf{h}^{(L_\mathrm{DM})},\, \mathbf{h}^{(L_\mathrm{SV})}$
\State $s \gets |\mathbf{x}|$ \Comment{committed length}
\Statex
\While{$s < T$}
  \State \textbf{// ===== Stage 1: Draft =====}
  \State $\mathbf{d} \gets [\,]$;\quad $\mathbf{H}_\mathrm{DM} \gets [\,]$
  \For{$k = 0,\ldots,\bar{\gamma}{-}1$}
    \State $h \gets \mathcal{M}_{1{:}L_\mathrm{DM}}(y_{s+k},\, \mathrm{KV}_\mathcal{M})$;\quad append $h$ to $\mathbf{H}_\mathrm{DM}$
    \State $(o,\, \mathrm{KV}_{\mathcal{A}_\mathrm{DM}}) \gets \mathcal{A}_\mathrm{DM}(h,\, \mathrm{KV}_{\mathcal{A}_\mathrm{DM}})$
    \State $y_{s+k+1} \gets \arg\max \mathcal{H}(o)$;\quad append to $\mathbf{d}$
  \EndFor
  \Statex
  \State \textbf{// ===== Stage 2: Shallow Verification =====}
  \State $\mathbf{H}_\mathrm{SV} \gets \mathcal{M}_{L_\mathrm{DM}{+}1{:}L_\mathrm{SV}}(\mathbf{H}_\mathrm{DM},\, \mathrm{KV}_\mathcal{M})$ \Comment{$\bar{\gamma}$ positions in parallel}
  \State $(\mathbf{O}_\mathrm{SV},\, \mathrm{KV}_{\mathcal{A}_\mathrm{SV}}) \gets \mathcal{A}_\mathrm{SV}(\mathbf{H}_\mathrm{SV},\, \mathrm{KV}_{\mathcal{A}_\mathrm{SV}})$;\quad $\hat{\mathbf{c}} \gets \arg\max \mathcal{H}(\mathbf{O}_\mathrm{SV})$
  \State $\mathcal{R} \gets \{\,(i,\, \hat{c}_i)\;:\; \hat{c}_i \neq d_i,\; 0 \le i < \bar{\gamma}\,\}$ \Comment{positions where SV proposes a correction}
  \Statex
  \If{$\mathcal{R} = \emptyset$}
    \State $\mathcal{T} \gets \textsc{ChainTree}(\mathbf{d})$ \Comment{straight chain, no correction branches}
  \Else
    \State $\mathcal{T} \gets \textsc{BuildTree}(\text{main}=\mathbf{d},\, \text{corrections}=\mathcal{R})$
    \Statex \quad\;\, \Comment{main branch is unchanged; each $(i,\hat{c}_i)\in\mathcal{R}$ adds a side branch at position $i$}
    \State Re-forward correction tokens through layers $1{:}L_\mathrm{SV}$ under tree-masked attention
    \Statex \quad\;\, \Comment{reuses DRAM-resident DM and SV layers; \textbf{zero} additional flash transfer}
  \EndIf
  \Statex
  \State \textbf{// ===== Stage 3: Final Verification (stream layers $L_\mathrm{SV}{+}1{:}L_\mathrm{total}$ from flash) =====}
  \State $\mathbf{H}_\mathrm{tgt} \gets \mathcal{M}_{L_\mathrm{SV}{+}1{:}L_\mathrm{total}}\!\big(\mathcal{T}.\text{hidden\_states},\; \text{mask}{=}\mathcal{T}.\text{tree\_mask},\; \mathrm{KV}_\mathcal{M}\big)$
  \State $\mathbf{p}_\mathrm{tgt} \gets \mathrm{softmax}(\mathcal{H}(\mathbf{H}_\mathrm{tgt}))$
  \Statex
  \State \textbf{// ----- Walk the tree to find the longest accepted prefix -----}
  \State $a \gets 0$;\quad $\mathbf{y}_\mathrm{out} \gets [\,]$
  \For{$i = 0,\ldots,\bar{\gamma}{-}1$}
    \If{\textsc{Acc}$(d_i,\, \mathbf{p}_\mathrm{tgt}^{(i)})$}
      \State $a \gets a + 1$;\quad append $d_i$ to $\mathbf{y}_\mathrm{out}$
    \ElsIf{$(i,\, \hat{c}_i) \in \mathcal{R}$ \textbf{and} \textsc{Acc}$(\hat{c}_i,\, \mathbf{p}_\mathrm{tgt}^{(\mathrm{corr},\, i)})$}
      \State $a \gets a + 1$;\quad append $\hat{c}_i$ to $\mathbf{y}_\mathrm{out}$;\quad \textbf{break}
      \Statex \quad\quad\quad \Comment{correction branch terminates the cycle by construction}
    \Else
      \State \textbf{break}
    \EndIf
  \EndFor
  \State Append $\arg\max \mathbf{p}_\mathrm{tgt}^{(a)}$ to $\mathbf{y}_\mathrm{out}$ \Comment{bonus token from target's own next-token prediction}
  \Statex
  \State Commit $\mathbf{y}_\mathrm{out}$ to $\mathbf{y}$;\quad $s \gets s + |\mathbf{y}_\mathrm{out}|$
  \State Roll back $\mathrm{KV}_\mathcal{M},\, \mathrm{KV}_{\mathcal{A}_\mathrm{DM}},\, \mathrm{KV}_{\mathcal{A}_\mathrm{SV}}$ to length $s$ along the accepted path
\EndWhile
\State \Return $\mathbf{y}_{0{:}s}$
\end{algorithmic}
\end{algorithm}

\clearpage
\section{Reduced KL Loss Computation Process}\label{app:loss}

In this appendix we expand the construction of the \textbf{Reduced KL Loss} introduced in Section~\ref{methodology}, Equation~\ref{eq:topk_loss}. Recall the setup: we train the draft and shallow-verifier adapters by aligning their output distribution $q_{\text{draft}}(v \mid t)$ with the target-model distribution $p_{\text{target}}(v \mid t)$ across a set $M$ of unmasked token positions, where $v \in \{1,\ldots,V\}$ indexes vocabulary tokens and $t$ indexes positions. The standard full-vocabulary distillation loss in Equation~\ref{eq:full_loss},
\begin{equation*}
    \mathcal{L}_{\text{full}} = -\frac{1}{|M|} \sum_{t \in M} \sum_{v=1}^{V}
    p_{\text{target}}(v \mid t) \cdot \log q_{\text{draft}}(v \mid t),
\end{equation*}
spreads supervision over all $V$ tokens, but the vast majority of vocabulary mass at any position $t$ sits in a long tail of near-zero probabilities that rarely determines acceptance. To concentrate the limited capacity of our compact adapters on the tokens that actually matter, we restrict the loss to the top-$K$ tokens of $p_{\text{target}}(\cdot \mid t)$ at each position. The construction proceeds in three steps.

\paragraph{Step 1: Select the top-$K$ indices.}
At each unmasked position $t \in M$, identify the $K$ vocabulary tokens carrying the highest probability mass under the target distribution:
\begin{equation}
    \mathcal{T}_K(t) = \operatorname*{arg\,top\text{-}K}_{v \in \{1,\ldots,V\}}
    \; p_{\text{target}}(v \mid t).
    \label{eq:topk_select}
\end{equation}
$\mathcal{T}_K(t)$ is a position-dependent support of size $|\mathcal{T}_K(t)| = K$ that captures the tokens with non-negligible probability of being emitted by the target model. In speculative decoding, only tokens within this set can plausibly be accepted, so any supervision spent outside $\mathcal{T}_K(t)$ is effectively wasted.

\paragraph{Step 2: Renormalize over the focused support.}
The masses $\{p_{\text{target}}(v \mid t)\}_{v \in \mathcal{T}_K(t)}$ in general do not sum to $1$. We renormalize them onto $\mathcal{T}_K(t)$ to obtain a valid probability distribution $\tilde{p}_{\text{target}}(v \mid t)$:
\begin{equation}
    \tilde{p}_{\text{target}}(v \mid t) =
    \begin{cases}
        \dfrac{p_{\text{target}}(v \mid t)}
              {\displaystyle\sum_{v' \in \mathcal{T}_K(t)} p_{\text{target}}(v' \mid t)},
        & v \in \mathcal{T}_K(t), \\[10pt]
        0,
        & v \notin \mathcal{T}_K(t).
    \end{cases}
    \label{eq:renorm}
\end{equation}
Equivalently, with an indicator function $\mathbf{1}[\cdot]$:
\begin{equation}
    \tilde{p}_{\text{target}}(v \mid t) =
    \frac{p_{\text{target}}(v \mid t) \cdot \mathbf{1}[v \in \mathcal{T}_K(t)]}
         {\displaystyle\sum_{v' \in \mathcal{T}_K(t)} p_{\text{target}}(v' \mid t)}.
     \label{eq:renorm_indicator}
\end{equation}
By construction $\sum_{v=1}^{V} \tilde{p}_{\text{target}}(v \mid t) = 1$, so $\tilde{p}_{\text{target}}(\cdot \mid t)$ is a proper distribution supported on $\mathcal{T}_K(t)$ and zero elsewhere.

\paragraph{Step 3: Cross-entropy on the focused support.}
The Reduced KL Loss is the position-averaged cross-entropy between $\tilde{p}_{\text{target}}(\cdot \mid t)$ and the adapter distribution $q_{\text{draft}}(\cdot \mid t)$, evaluated only on $\mathcal{T}_K(t)$:
\begin{equation}
    \mathcal{L}_{\text{top-}K} = -\frac{1}{|M|} \sum_{t \in M} \sum_{v \in \mathcal{T}_K(t)}
    \tilde{p}_{\text{target}}(v \mid t) \cdot \log q_{\text{draft}}(v \mid t).
    \label{eq:topk_loss2}
\end{equation}
Because $\tilde{p}_{\text{target}}(v \mid t) = 0$ for $v \notin \mathcal{T}_K(t)$, the inner sum is mathematically equivalent to summing over the entire vocabulary; in practice we materialize only the $K$ active terms per position to avoid the $V{-}K$ vanishing contributions. Up to a constant entropy term $H\!\left(\tilde{p}_{\text{target}}(\cdot \mid t)\right)$ that does not depend on the adapter parameters, $\mathcal{L}_{\text{top-}K}$ coincides with the position-averaged KL divergence
\begin{equation*}
    \frac{1}{|M|} \sum_{t \in M} \mathrm{KL}\!\left(\tilde{p}_{\text{target}}(\cdot \mid t)\;\|\;q_{\text{draft}}(\cdot \mid t)\right),
\end{equation*}
which is why we call it the \emph{Reduced} KL Loss: it is a KL divergence taken over a reduced support $\mathcal{T}_K(t)$ rather than the full vocabulary $V$.

\paragraph{Effect on adapter training.}
Compared with $\mathcal{L}_{\text{full}}$, restricting supervision to $\mathcal{T}_K(t)$ has two effects. First, it concentrates gradient signal on the candidates that drive acceptance, since tokens outside $\mathcal{T}_K(t)$ have $p_{\text{target}}$ values too small to be accepted regardless of how $q_{\text{draft}}$ allocates mass to them. Second, it stabilizes training under our compact draft and shallow-verifier sub-networks: the long tail of near-zero target probabilities no longer consumes adapter capacity. This is essential in our memory-limited setting, where the draft and SV sub-networks must remain small enough to fit in the DRAM budget while still producing high-acceptance candidates.



\end{document}